\documentclass[article, nojss]{jss}\usepackage[]{graphicx}\usepackage{xcolor}
\makeatletter
\def\maxwidth{ %
  \ifdim\Gin@nat@width>\linewidth
    \linewidth
  \else
    \Gin@nat@width
  \fi
}
\makeatother

\usepackage{Sweave}

\usepackage{thumbpdf,lmodern}
\graphicspath{{Figures/}}
\usepackage[utf8]{inputenc}
\usepackage{amsmath}
\usepackage{amsfonts}
\usepackage{a4wide}
\usepackage{multirow}

\usepackage{tikz}
\usepackage{comment}
\usepackage{lscape}
\usepackage{rotating}

\definecolor{darkgray}{gray}{0.3} 
\definecolor{darkgreen}{cmyk}{0.34,0.0,0.69,0.56} 
\definecolor{darkred}{cmyk}{0.0,0.65,0.92,0.31} 
\definecolor{marille}{cmyk}{0.0,0.4,0.67,0.0} 
\definecolor{Red}{rgb}{0.5,0,0}
\definecolor{Blue}{rgb}{0,0,0.5}

\newcommand{\R}{\mathbb{R}}
\newcommand{\K}{\mathcal{K}}

\newcommand{\norm}[1]{\left\lVert#1\right\rVert}

\newcommand{\comnt}[2][COMMENT]{}

\newcommand{\CM}[0]{\checkmark}

\author{Benjamin Schwendinger\\TU Wien
   \And \hspace{3em}Florian Schwendinger\\\hspace{3em}University of Klagenfurt
   \And Laura Vana\\TU Wien
}

\title{Holistic Generalized Linear Models}

\Plainauthor{Benjamin Schwendinger, Florian Schwendinger, Laura Vana}
\Plaintitle{Holistic Generalized Linear Models}
\Shorttitle{Holistic GLMs}

\Abstract{ 
    Holistic linear regression extends the classical best subset selection problem
    by adding additional constraints designed to improve the model quality.
    These constraints include sparsity-inducing constraints, sign-coherence
    constraints and linear constraints.
    The \proglang{R} package \pkg{holiglm} provides functionality to model and fit
    holistic generalized linear models.
    By making use of state-of-the-art conic mixed-integer
    solvers, the package can reliably solve GLMs for Gaussian,
    binomial and Poisson responses with a multitude of holistic
    constraints. The high-level interface simplifies the constraint
    specification and can be used as a drop-in replacement for
    the \code{stats::glm()} function.
}

\Keywords{Algorithmic regression, best subset selection, 
conic programming, holistic constraints, optimization, \proglang{R}}
\Plainkeywords{Generalized linear models, holistic regression,
    algorithmic regression, best subset selection,
    conic programming, holistic constraints, optimization, R} 

\Address{
  Benjamin Schwendinger\\
  Institute of Computer Technology\\
  TU Wien\\
  Gu{\ss}hausstra{\ss}e 27-29\\
  1040 Vienna, Austria\\
  E-mail: \email{benjamin.schwendinger@hotmail.com}\\ \\
  Florian Schwendinger\\
  Department of Statistics\\
  University of Klagenfurt\\
  Universit\"atsstra{\ss}e 65-67\\
  9020 Klagenfurt, Austria\\
  E-mail: \email{FlorianSchwendinger@gmx.at}\\ \\
  Laura Vana \\
  CSTAT -- Computational Statistics \\
  Institute of Statistics and Mathematical Methods in Economics\\
  TU Wien\\
  Wiedner Hauptstra{\ss}e 7 \\
  1040 Vienna, Austria\\
  E-mail: \email{laura.vana.guer@tuwien.ac.at}
}
\IfFileExists{upquote.sty}{\usepackage{upquote}}{}
\begin{document}
%\SweaveOpts{concordance=TRUE, prefix.string=Figures/jss}

% ------------------------------------------------------------------------------
%
% Introduction
%
% ------------------------------------------------------------------------------
\section{Introduction}\label{sec:introduction}
Selecting a sensible model from the set of all possible models is an important 
but typically time-consuming task in the data analytic process. 
To simplify this process, \cite{algoReg:Bertsimas:2015, holiReg:Bertsimas2020} 
introduce the holistic linear model (HLM). 
The HLM is a constrained linear regression model in which the constraints aim
to automate the model selection process. In particular, this can be achieved
by utilizing quadratic mixed-integer optimization, where the integer constraints
are used to place cardinality constraints on the linear regression model.

Cardinality constraints are used to introduce sparsity in the statistical
model, a desirable property that leads to more interpretability, increase in 
computational efficiency (especially in high-dimensional settings) and  
reduction in the variance of the estimates 
(at the cost of introducing some bias).
Placing a cardinality constraint on the total 
number of variables allowed in the final model leads to the classical best 
subset selection
problem \citep{subsetSelection:Miller:2002} which, given a response vector 
$y \in \mathbb{R}^n$, a predictor matrix $X \in \mathbb{R}^{n\times p}$ and a 
subset size $k$ between 0 and $\min\{n, p\}$, finds the subset of at most $k$ predictors that 
produces the best fit in terms of squared error, solving the nonconvex problem
\begin{equation*} % \label{eqn:subset}
    \underset{\beta}{\text{minimize }} ~~ \frac{1}{2} \norm{y - X \beta }_2^2
    ~~ \textrm{subject to} ~~
    \sum_{i=1}^p \mathbb{I}_{\{\beta_i \neq 0\}} \leq k.
\end{equation*}
Further cardinality constraints placed on user-defined groups of predictors 
can be used to e.g. limit the pairwise multicollinearity,
select the best (non-linear) transformation for the predictors, include expert
knowledge in the model estimation (such as forcing specific predictors
to stay in the model) or ensure that certain sets of predictors are jointly 
included (excluded) in (from) the model.

In addition to cardinality constraints, upper or lower bounds as well as 
linear constraints on the coefficients are also relevant for improving 
interpretability along with introducing sparsity 
\citep[e.g. see][who introduce non-negativity constraints in linear regression and
show performance similar to other sparsity inducing methods]{nnls:Slawski2013}.

In this paper, we introduce the \pkg{holiglm} package, an \proglang{R} package for 
formulating and fitting holistic generalized linear models (HGLMs). 
The package supports holistic constraints such as mixed-integer-type constraints 
related to sparsity (e.g. limits on the number of covariates to be included in 
the model, group sparsity constraints), linear constraints and constraints 
related to multicollinearity. 
The contribution of the paper is three-fold. 
First, to the best of our 
knowledge, we are the first to suggest the use of conic optimization to extend the 
results presented for linear regression by 
\cite{algoReg:Bertsimas:2015, holiReg:Bertsimas2020}
to the class of generalized linear models and to formulate the most common GLMs as 
conic optimization problems. Secondly, we survey 
the literature related to holistic linear regression and, more generally, to 
constrained regression and provide an extensive survey of what should be considered
as holistic constraints. Finally, we provide a ready-to-use implementation
of holistic GLMs in the package \pkg{holiglm}. We exemplify the use of the package 
for a variety of statistical problems with the hope that this will encourage 
the statistical community in further exploiting the recent advances in conic mixed-integer 
optimization. 

The \pkg{holiglm} package provides a flexible infrastructure for automatically 
translating constrained generalized linear models into conic optimization problems.
The optimization problems are solved by utilizing the \proglang{R} optimization
infrastructure package \pkg{ROI} \citep{roi:theussl:2020}.
Additionally, a high-level interface, which can be used as a
drop-in replacement for the \code{stats::glm()} function, is provided.
 Using \pkg{ROI}
makes it possible for the user to choose between a wide range of commercial
and open source optimization solvers. 
With recent advancements in conic optimization, it is now possible to routinely
solve conic problems with a large number of variables and/or constraints to proven optimality.
Using conic optimization instead of iteratively reweighted least squares (IRLS)
has the advantages that no starting values are needed, the results are more
reliable and the solvers are designed to handle different
types of constraints.
These advantages come at the cost of a longer runtime; however, as 
shown by \cite{log-bin:schwendinger:2021} for some GLMs the speed of the conic 
formulation is similar to the IRLS implementation.

At the time of writing, there exists no ready-to-use 
package or library for fitting HGLMs or HLMs. However, for 
\proglang{R}~\citep{R:R-Core} several packages allow the estimation 
of (generalized) linear models under linear constraints.
The \pkg{CMLS} \citep{pkg:cmls:2018} package can fit linear regression models under linear constraints
by translating the problem into a linear constrained quadratic
optimization problem, which is solved by \pkg{quadprog} \citep{pkg:quadprog:2019}.
Package \pkg{colf}~\citep{pkg:colf:2017} can fit linear
regression models with lower and upper bounds on the coefficients.
The \pkg{glmc} \citep{pkg:glmc:2006} package can fit generalized linear models 
while imposing linear constraints on the parameters.
Similarly, package \pkg{restriktor} \citep{pkg:restriktor2022} provides functionalities for 
estimation, testing and evaluation of linear equality and inequality constraints
about parameters and effects for generalized linear models, where the constraints 
can be specified by a text-based description. 
Finally, \pkg{ConsReg} \citep{pkg:ConsReg:2022} provides a similar functionality
to \pkg{restriktor}.

This paper is structured as follows:
Section~\ref{sec:glm} introduces the GLM class and 
presents the representation of specific family-link combinations
as conic optimization models.
In Section~\ref{sec:holistic_constraints}, we present a list of
holistic constraints implemented in the package.
Section~\ref{sec:package} introduces package \pkg{holiglm} for \proglang{R} and 
Section~\ref{sec:use_cases} shows the usage of the package in two applications:
fairness constraints in logistic regression and model selection in log-binomial regression.
Section~\ref{sec:conclusions} concludes.

% ------------------------------------------------------------------------------
%
% GLMs
%
% ------------------------------------------------------------------------------
\section{Generalized linear models}\label{sec:glm}
A generalized linear model (GLM) is a model for a response variable whose 
conditional distribution belongs to a one-dimensional exponential family.
A GLM consists of a linear predictor
$$
\eta_i = \beta_0 + \beta_1x_{1i} +\ldots +\beta_p x_{pi}=\boldsymbol x_i^\top\boldsymbol \beta
$$
and two functions, namely a twice continuously differentiable and invertible 
link function $g$ that describes how the mean, 
$\E(Y_i) = \mu_i$, depends on the linear predictor (one has $g(\mu_i) = \eta_i$), and a 
variance function $V$ that describes how the variance depends on the 
mean $\VAR(Y_i) = \phi V(\mu)$ (where the dispersion parameter $\phi > 0$ is a constant).

The most common distributions for the response $Y_i$ that can be accommodated in 
the GLM setting include the normal, binomial or Poisson.
The density of members of the exponential family can be written as:
$$
f(y; \lambda, \phi)=\mathrm{exp}
  \left\{\frac{y\lambda-b(\lambda)}{\phi}+c(y,\phi)\right\}\text{‚}
$$
where $\lambda$ is called the canonical or natural parameter, $b(\cdot)$ and 
$c(\cdot)$ are known real-valued
measurable functions that vary from one exponential family to another.

It can be shown that
$\E(y)=b^\prime(\lambda)$ and $\VAR(y)=\phi b^{\prime\prime}(\lambda)$ holds.
Note that $g(b^\prime(\lambda))=\eta_i$.
We can denote $h=(b^\prime)^{-1} \circ g^{-1}$ so 
$\lambda(\boldsymbol\beta)=h(\eta)$.
In case $h$ is the identity, the link $g$ is called canonical, i.e., 
$b'(\cdot)$ is the inverse of the canonical link function.

The parameters $\boldsymbol\beta$ can be 
estimated by maximizing the likelihood function corresponding to the different 
types of response. Using the distribution of the exponential family,
the log-likelihood for a sample $\boldsymbol y = (y_1, \ldots, y_n)^\top$ is given by:
\begin{align}\label{eq:glmexpfam}
\log \mathcal{L}(\boldsymbol\beta; \boldsymbol y) = \sum_{i=1}^{n}
  \frac{y_i\lambda_i(\boldsymbol\beta) - 
  b(\lambda_i(\boldsymbol\beta))}{\phi_i} + c(y_i, \phi_i).
\end{align}
where $\phi_i=\phi/a_i$ for $a_i$ are user-defined observation weights. 

In this paper we leverage the fact that, for the most common family-link combinations,
the maximization of the objective function in Equation~\ref{eq:glmexpfam} can be
reformulated as a conic optimization problem.
A conic optimization problem is designed to model convex problems
by optimizing a linear objective function 
over the intersection of an affine hyperplane and a nonempty closed convex cone.
For GLMs, the types of cones relevant for maximum likelihood estimation include
% 1)
the exponential cone $\mathcal K_\text{exp}$, which can be used to model a variety of objectives and 
constraints involving exponentials and logarithms,
% 2)
and
the second-order cone $\K_\text{soc}$, which can be 
used to model problems involving -- directly or indirectly -- quadratic terms. 
% 3)
Table~\ref{tab:glms} presents the log-likelihood functions for the 
family-link combinations implemented in package \pkg{holiglm}, and
their reformulation as conic problems. 
More details on these reformulations can be found in  
Appendix~\ref{sec:formulations}.

Inference is based on maximum likelihood, unless the 
holistic constraints presented in Section~\ref{sec:holistic_constraints} are 
binding. In the case of binding constraints,  we employ the constrained maximum 
likelihood framework in \cite{schoenberg1997constrained} and 
apply a correction to the standard errors 
of the coefficients.

\begin{table}[t!]
\begin{small}
\begin{tabular}{l l c c}
  \hline
  % --------------------------------------------------------
  Family-
  & $\lambda_i(\boldsymbol\beta)$, $b_i(\boldsymbol\beta)$
  & Objective & Conic \\
  % --------------------------------------------------------
  link& &  &
  constraint\\
  % --------------------------------------------------------
  \hline
  % --------------------------------------------------------
  \multicolumn{4}{l}{\textbf{Gaussian}} \\
  % --------------------------------------------------------
  %\hline
  % --------------------------------------------------------
  Identity$^a$ 
  & $\lambda_i(\boldsymbol\beta)=\boldsymbol x_i^\top\boldsymbol\beta$ , 
  & $-\zeta$
  & $( \zeta + 1,
  \zeta - 1,
  2( y_1 -  \boldsymbol x_1^\top \boldsymbol\beta ),$
  \\
  &$b_i(\boldsymbol\beta)\,\,\,=(\boldsymbol x_i^\top\boldsymbol\beta)^2/2$ &
  & $\ldots, 2( y_n -  \boldsymbol x_n^\top\boldsymbol \beta )) \in \K_\text{soc}^{n+2}$ \\
  % --------------------------------------------------------
  % \hline
  % --------------------------------------------------------
  \multicolumn{4}{l}{\textbf{Binomial}} \\
  % --------------------------------------------------------
  %\hline
  % --------------------------------------------------------
  Logit$^b$ \& 
    & $\lambda_i(\boldsymbol\beta)=\boldsymbol x_i^\top\boldsymbol\beta$
  % & $\log(1+\exp(\boldsymbol x_i^\top\boldsymbol\beta))$
  & $\sum\limits_{i=1}^ny_i\boldsymbol x_i^\top\boldsymbol\beta-\delta_i$ &
   $(\delta_i, 1, \gamma_i+1) \in \mathcal{K}_\text{exp}$ \\
  % --------------------------------------------------------
  Probit& $b_i(\boldsymbol\beta)\,\,\,=\log(1+\exp(\boldsymbol x_i^\top\boldsymbol\beta))$ &  & $(\boldsymbol x_i^\top\boldsymbol\beta,1,\gamma_i)\in \mathcal{K}_\text{exp}$ \\[6pt]
  % --------------------------------------------------------
  % --------------------------------------------------------
  Log$^b$  
  & $\lambda_i(\boldsymbol\beta)=\boldsymbol x_i^\top\boldsymbol\beta-\log(1-\exp(\boldsymbol x_i^\top\boldsymbol\beta))$
 % & $-\log\left(1-\exp(\boldsymbol x_i^\top\boldsymbol\beta)\right)$ 
  & $\sum\limits_{i=1}^ny_i\boldsymbol x_i^\top\boldsymbol\beta + (1-y_i)\delta_i$ & $\boldsymbol x_{i}^\top \boldsymbol\beta \leq 0$ \\
  % % --------------------------------------------------------
   & $b_i(\boldsymbol\beta)\,\,\,=-\log\left(1-\exp(\boldsymbol x_i^\top\boldsymbol\beta)\right)$  &
   & $(\delta_i,1,1-\gamma_i) \in \mathcal{K}_\text{exp}$ \\
  % % --------------------------------------------------------
   & & & $(\boldsymbol x_i^\top\boldsymbol\beta,1,\gamma_i)\in \mathcal{K}_\text{exp}$ \\
  % % --------------------------------------------------------
  % % Cloglog$^b$ & $\log(1-\exp(-\exp(\boldsymbol x_i^\top\boldsymbol\beta)))+$ &$\exp(\boldsymbol x_i^\top\boldsymbol\beta)$   & $y_i\delta_i - (1-y_i)\zeta_i $ & $\delta_i \geq \log(1-\exp(-\exp(\boldsymbol x_i^\top\boldsymbol\beta)))\iff$ \\
  % %  & $\quad+\exp(\boldsymbol x_i^\top\boldsymbol\beta)$  & & & $(\delta_i,1,1-\gamma_i) \in \mathcal{K}_\text{exp}$  \\
  % %  &  & & &$(\gamma_i, 1, -\zeta_i) \in \mathcal{K}_\text{exp}$  \\
  % %  & & & &$\zeta_i \leq \exp(\boldsymbol x_i^\top\boldsymbol\beta)\iff$  \\
  % %  & & & &$(\boldsymbol x_i^\top\boldsymbol\beta,1,\zeta_i) \in \mathcal{K}_\text{exp}$  \\
  % % --------------------------------------------------------
  % \hline
  % % --------------------------------------------------------
  \multicolumn{4}{l}{\textbf{Poisson}} \\
  % % --------------------------------------------------------
  % \hline
  % % --------------------------------------------------------
   Log$^b$
   & $\lambda_i(\boldsymbol\beta)=\boldsymbol x_i^\top\boldsymbol\beta$
  % & $\exp(\boldsymbol x_i^\top\boldsymbol\beta)$
   & $\sum\limits_{i=1}^n y_i\boldsymbol x_i^\top\boldsymbol\beta - \delta_i$
   & $(\boldsymbol x_i^\top\boldsymbol\beta,1,\delta_i)\in \mathcal{K}_\text{exp}$ \\
   & $b_i(\boldsymbol\beta)\,\,\,=\exp(\boldsymbol x_i^\top\boldsymbol\beta)$ & &   \\[6pt]
  % % --------------------------------------------------------
    Identity$^b$
   & $\lambda_i(\boldsymbol\beta)=\log(\boldsymbol x_i^\top\boldsymbol\beta)$
  % & $\boldsymbol x_i^\top\boldsymbol\beta$
   & $\sum\limits_{i=1}^ny_i\delta_i - \boldsymbol x_i^\top\boldsymbol\beta$
   & $(\delta_i, 1, \boldsymbol x_i^\top\boldsymbol\beta)\in \mathcal{K}_\text{exp}$ \\
   & $b_i(\boldsymbol\beta)\,\,\,=\boldsymbol x_i^\top\boldsymbol\beta$ &
   & $\boldsymbol x_{i}^\top \boldsymbol\beta \geq 0$ \\[6pt]
  % % --------------------------------------------------------
    Sq.root$^{ab}$
    & $\lambda_i(\boldsymbol\beta)=2\log(\boldsymbol x_i^\top\boldsymbol\beta)$
  % & $(\boldsymbol x_i^\top\boldsymbol\beta)^2$
    & $ \sum\limits_{i=1}^n 2y_i\delta_i - \zeta$&$(\delta_i, 1, \boldsymbol x_i^\top\boldsymbol\beta) \in \mathcal{K}_\text{exp}$ \\
  % % --------------------------------------------------------
    &$b_i(\boldsymbol\beta)\,\,\,=(\boldsymbol x_i^\top\boldsymbol\beta)^2$ &
    &$(\zeta + 1,
    \zeta - 1,
    2 \boldsymbol x_{1}^\top \boldsymbol\beta,$  \\
    & & & $\ldots,
    2 \boldsymbol x_{n}^\top \boldsymbol\beta
   ) \in \mathcal{K}_\textrm{soc}^{n+2}$ \\
\hline
\end{tabular}
\end{small}
\caption{GLMs as exponential families with conic reformulations.
The objective functions to be maximized are proportional to 
$\propto \sum_{i=1}^ny_i\lambda_i(\boldsymbol\beta)-b_i(\boldsymbol\beta)$.
%All expressions are for one observation $i$. 
The conic constraints with subscript~$i$ should hold for all $i=1,\ldots,n$.
Superscript $^a$ marks family-link combinations with an objective
modeled by second-order cones. 
Superscript $^b$ marks family-link combinations with an exponential
cone representation. 
Note that for the probit link an approximation 
is used by scaling the coefficients obtained from the logit link by $\sqrt{\pi/8}$
which is equivalent to using a simple approximation \citep{Page1977Approx} 
for the standard normal distribution.}\label{tab:glms}
\end{table}
%\end{table}

% ------------------------------------------------------------------------------
%
% Holistic constraints}
%
% ------------------------------------------------------------------------------
\section{Holistic constraints}\label{sec:holistic_constraints}
\cite{algoReg:Bertsimas:2015} and \cite{holiReg:Bertsimas2020} introduced a set
of constraints designed to improve the quality of linear regression models.
These constraints are crafted based on a survey of modeling recommendations from 
statistical textbooks and articles. 
In this section we survey different constraints suggested 
by the literature and provide an extensive overview which includes
but is not limited to constraints in 
\cite{algoReg:Bertsimas:2015} and \cite{holiReg:Bertsimas2020}. Moreover, we provide their 
formulations as optimization constraints.

Most of the constraints introduced in \pkg{holiglm} are cardinality constraints.
For modeling cardinality constraints we need $p$ binary variables:
\begin{equation}
  z_j \in \{0, 1\}, ~~~ j = 1, \dots, p.
\end{equation}
These binary variables are only added to the model if needed.
Here, the binary variable $z_j$ represents the selection of variable 
$X_j$, hence, $z_j = 0$ implies $\beta_j = 0$. Problems which involve
such binary variables are called mixed-integer optimization problems.
This type of cardinality constraints can be modeled with a so-called big-$M$
constraint.
\begin{equation}
  -M z_j \leq \beta_j \leq M z_j, ~~~ j = 1, \dots, p,
\end{equation}
where $M$ is a positive constant.
For the big-$M$ constraint it is important to choose a good constant $M$.
A good $M$ can be characterized by two properties:
\begin{itemize}
\item it is chosen big enough, that it does not impact the magnitude of the
      parameter $\beta_i$.
\item it is chosen as small as possible to improve the speed of the underlying 
      optimization solver and ensure stability.
\end{itemize}

\subsection{Global sparsity}
The global sparsity constraint can be used to model the classical best subset 
selection problem \citep{subsetSelection:Miller:2002}.
The best subset selection problem is a combinatorial optimization problem designed
to select a predefined number of $k_\text{max}$ covariates out of all available covariates.
This can be accomplished by maximizing the likelihood
function while restricting the number of covariates via integer constraints:
$$
  \sum_{j=1}^p z_j \leq k_\text{max}, ~~~ \text{ where } k_\text{max}\leq p, k_\text{max}\in \mathbb{N}.
$$

\subsection{Group sparsity}
Contrary to global sparsity constraints, group sparsity constraints do not 
restrain all covariates but only specific local groups of covariates.
Group sparsity constraints can again be subdivided into excluding and including 
constraints.
\begin{itemize}
\item excluding constraints enforce a cardinality constraint on a subset of covariates;
\item including constraints enforce that either all covariates of a group are selected or none.
\end{itemize}
\cite{algoReg:Bertsimas:2015} suggest to use excluding constraints for modeling 
pairwise multicollinearity and selecting the best (non-)linear transformation.
Including constraints can be used for modeling categorical variables via one-hot
encoding or splines via spline basis functions. In the following we present
the formulation of the excluding group sparsity constraints for restricting
pairwise multicollinearity and for selecting the best variable transformation.

\subsubsection{Limited pairwise multicollinearity}
A desired property for a good model is that the covariates exhibit no 
multicollinearity. Collinear covariates can cause numerical problems during the 
estimation, less interpretable coefficients and misleading standard errors.

The group sparsity constraint can be used to restrict pairwise multicollinearity 
by excluding pairs of covariates that exhibit strong pairwise correlation.

Using the previously introduced variables $z$, the pairwise 
multicollinearity can be formulated as,
$$
  z_j + z_k \leq 1 ~~~ \forall (j,k) \in \mathcal{PC},
$$
here $\mathcal{PC} = \{(j,k) ~ | ~ \rho_\text{max} < |\rho(j,k)|\}$ denotes the set 
of pairs of highly correlated covariates.
The general rule of thumb is excluding pairs of covariates where the absolute 
value of the pairwise correlation exceeds the threshold $\rho_\text{max} = 0.7$, 
but other -- less as well as more restrictive -- choices are also advocated for 
in the literature 
\citep[see][for a review]{collinearity:dormann:2013}.

\subsubsection{Non-linear transformations}
In cases where the true relationship between the response and a covariate is non-linear,
using an appropriate transformation can improve model quality.
Commonly non-linear transformations like $\log(x)$, $\sqrt{x}$ or $x^2$ are used.
When using non-linear transformations, it is often desirable to only include either
the non-transformed variable or only one of the transformed variables.
A group sparsity constraint of the form
$$
  \sum_{j \in \mathcal{NL}} z_j \leq 1,
$$
can be used to select at most one of the transformations.
Here $\mathcal{NL}$ denotes the set of indices of applied transformations, including the 
identity mapping, on a certain covariate.

\subsubsection{Including group constraints}
In cases where all covariates within a specified group $\mathcal{G}$ 
should be included in the model (as in the case of dummy-encoded categorical variables), 
the including group sparsity constraint (also referred here as in-out constraint)
is translated to equality of all corresponding $z_i$'s variables: 
$$
z_j=z_k \quad \forall (j,k)\in \mathcal{G}
$$

\subsection{Modeler expertise}
\cite{algoReg:Bertsimas:2015} point out that, in some situations, the modeler 
knows the true importance of a particular covariate, be it some business 
requirement or other more profound expert knowledge.
In these cases, the modeler can choose to force the inclusion of a particular 
covariate. This might be necessary when the automated covariate selection process
decides to discard the covariate due to other global or group sparsity constraints.
To achieve this inclusion, the following constraint can be used:
$$
z_j = 1.
$$

\subsection{Bounded domains for covariates}
In certain applications, it is of interest to set constraints (bounds)
on the value of particular regression coefficients. For example, \cite{McDonald1990}
consider the problem of finding maximum likelihood estimates of a generalized 
linear model when some or all regression parameters are constrained to be 
non-negative and motivate their approach with applications in cancer death prediction
and demography. Similarly,
\cite{nnls:Slawski2013} propose the use of non-negative least squares in
high-dimensional linear models. They show that this can have similar effects 
to explicit regularization like LASSO \citep{tibshirani:lasso:1996} or ridge 
regression \citep{hoerl:ridge:1970} regarding predictive power and sparsity 
when mild conditions on the design matrix are met.
Moreover, there exist problems where the sign of the effect of a covariate is known
beforehand. This allows the restriction of coefficient domains to only 
non-negative or only non-positive values \citep{interpretability:carrizosa:2020}.
In our conic model approach, for a coefficient $\beta_j$, we can add
bounds (sometimes also called box constraints) $l_j$ and $u_j$ by the following
constraint:
$$
  l_j \leq \beta_j \leq u_j.
$$
\subsection{Linear constraints} \label{con:linear_constraint}

As \cite{ls_lawson_1995} point out, it might also be desirable to enforce linear 
constraints of the type $L\boldsymbol\beta\leq c$ for certain covariates
(typical constraints include $\beta_j + \beta_k \leq \beta_l$, 
$\beta_j \leq \beta_k$ or $\beta_j = \beta_k = \beta_l$). 
Such constraints might arise in applied 
mathematics, physics and economics and usually convey additional information 
about a problem. 
Adding additional convex constraints to a problem will 
only tighten the set of solutions. One still has to be careful not to add 
constraints that make the underlying optimization problem infeasible.
If any infeasibility arises, the underlying optimization solver will detect it.

Note that such linear constraints can also be imposed on the binary variables
$z_i$. Said constraints can be used to model if-then types of relations. As an
example, consider a constraint of the type ``if $X_j$ is included then $X_k$
should also be included in the model''. This would translate to the
constraint $z_j\leq z_k$.  Note also that the global
and group sparsity constraints introduced above are special cases of linear 
constraints on the binary variables.

\subsection{Sign coherence constraint}
Sign coherence can be a desirable model property which improves 
interpretability, especially in the presence of highly correlated covariates.
\citet{interpretability:carrizosa:2020} suggest using a sign coherence constraint 
instead of the group sparsity constraint to restrict pairwise multicollinearity.
They argue that enforcing sign coherence (forcing strongly positive
correlated variables to have the same sign and strongly negative correlated 
variables to have opposite signs) is less restrictive than using a group sparsity 
constraint, since it allows strongly correlated variables to be jointly in the model
without lowering interpretability due to inconsistent signs. Without this 
restriction, inconsistent signs of strongly correlated variables are often 
caused by compensating coefficients \citep{multivariate:hahs:2016}.
Given the set $\mathcal{PC}$ of pairs of indices of highly correlated covariates, 
we can enforce equal signs of their respective coefficients by adding the constraints
$$
  -M(1-u_{jk}) \leq \beta_j, \beta_k \leq M u_{jk}, ~~~ \forall (j,k) \in \mathcal{PC},
$$
where again $M$ is a large enough constant and $u_{jk}$ is a newly introduced binary variable.
One can see that for $u_{jk}=0$ we have $-M \leq \beta_j, \beta_k \leq 0$, while for $u_{jk}=1$,
we have $0 \leq \beta_j, \beta_k \leq M$ for an arbitrary pair $(j,k)$ in $\mathcal{PC}$.
% ------------------------------------------------------------------------------
%
%
% Package
%
%
% ------------------------------------------------------------------------------
\section[The holiglm package]{The \pkg{holiglm} package}\label{sec:package}
Package \pkg{holiglm} allows to reliably and conveniently fit generalized linear
models under constraints.
We aimed to allow for as many family-link combinations and constraints 
as possible without reducing the reliability of the solution. To accomplish 
these goals, the package uses state-of-the-art mixed-integer (conic) solvers. 
Using conic optimization, we can reliably solve convex non-linear mixed-integer 
problems, given that there exists a combination of cones that
can express the non-linear problem at hand. Luckily, the (log-)likelihood of most
GLMs can be expressed by combinations of the linear (non-negative) cone, 
the second-order cone and the exponential cone. 
Table~\ref{tab:family_link_solver_overview} gives an overview on the cones used 
to express specific GLMs. 
Additionally, Table~\ref{tab:solver_overview} provides a list of the
solvers available from \pkg{ROI} which allow for mixed-integer 
constraints. Information on the different types of cones supported by these solvers
is also provided.
All the solvers listed in Table~\ref{tab:solver_overview}
internally aim to prove the optimality of the solution by checking criteria 
based on the Karush-Kuhn-Tucker optimality conditions. Consequently, if these 
solvers signal optimality, the user can be certain that the maximum likelihood
estimate (MLE) was obtained, 
except for the special case where the MLE does not exist. In theory, these 
solvers should be able to provide a certificate of unboundedness if the MLE
does not exist; however, practically this is often not the case for the exponential cone.
Fortunately, package \pkg{detectseparation}~\citep{pkg:detectseparation:2022} 
can be used to verify the existence of the MLE before the estimation.
More information on conic optimization in general can be found in 
\cite{Boyd+Vandenberghe:2004}.
\begin{table}[t!]
\centering
\begin{tabular}{@{}l|ccc@{}}
 \hline
 Link / Family
    & Gaussian
    & Binomial
    & Poisson
    \\
    \hline
  Identity
    & Q | SOC &  & LIN \& EXP
    \\
  Log
    &  & LIN \& EXP & EXP
    \\
  Logit
    &  & EXP
    \\
  Probit
    &  & EXP
    \\
  Square root
    &  &  & SOC \& EXP
    \\
  \hline
  \end{tabular}
  \caption[Overview on the cones needed to express specific GLMs.]{
    Overview of the cones needed to express specific GLMs. Here, Q | SOC indicates that
    either a quadratic solver or a solver supporting the second-order cone can be used.
    SOC \& EXP indicates that the exponential cone and second-order cone 
    are used to model the Poisson model with power link.  
    Note that for the probit link an approximation is used by scaling the 
    coefficients obtained from the logit link by $\sqrt{\pi/8}$ 
    \citep{Page1977Approx}.}
  \label{tab:family_link_solver_overview}
\end{table}

\begin{table}[t!]
\centering
\begin{tabular}{@{}lll|cccc@{}}
\hline
Solver & License & Package                 & Quadratic Objective & LIN & SOC & EXP \\
\hline
CPLEX  & Proprietary & \pkg{ROI.plugin.cplex}  & \CM \\
ECOS   & GPL-3       & \pkg{ROI.plugin.ecos}   &     & \CM & \CM & \CM \\
GUROBI & Proprietary & \pkg{ROI.plugin.gurobi} & \CM \\
MOSEK  & Proprietary & \pkg{ROI.plugin.mosek}  & \CM & \CM & \CM & \CM \\
       & Mixed       & \pkg{ROI.plugin.neos}   & \CM \\
\hline
\end{tabular}
\caption[Overview of the solvers implemented in ROI.]{Overview of the solvers 
   implemented as plugins in \pkg{ROI} that can handle mixed-integer constraints
   in combination with either 
   a quadratic objective with linear constraints or
   a linear objective with linear, second-order and exponential conic constraints.}
\label{tab:solver_overview}
\end{table}

In \proglang{R}, the packages \pkg{ROI} and \pkg{CVXR}~\citep{cvxr:2020} provide
access to multiple conic solvers. The \pkg{CVXR} package provides a
domain-specific language (DSL) which allows to formulate the optimization 
problems close to their mathematical formulation.
To accomplish this, it first translates the problem into an abstract syntax 
tree and uses disciplined convex programming (DCP) \citep{CVX:grant:2006} to verify 
that the problems are indeed convex. After the convexity is verified, the problem 
is transformed into its matrix representation and dispatched to the selected solver. 
Here it is important to note that the DCP rules are only sufficient for convexity, 
meaning that even if the DCP rules can not verify convexity, the problem can still  be convex. Based on these properties, from our perspective the \pkg{CVXR}
package is especially well-suited for users unfamiliar with conic optimization 
or fast prototyping. 
The \pkg{ROI} package, on the other hand, offers a unified solver access to many
solvers, with a  simple modeling language designed for users already familiar 
with \proglang{R}. For the \pkg{holiglm} package, we choose to rely on the infrastructure of package \pkg{ROI}  since we already know the convexity 
properties of the likelihood functions and we want to be able to control the transformation 
of the likelihood into the optimization problem. Additionally, \pkg{ROI} has fewer 
dependencies and, last but not least, the authors are familiar with package \pkg{ROI}.
\begin{Schunk}
\begin{Sinput}
R> library("holiglm")
\end{Sinput}
\end{Schunk}

%
% Model fitting
%
\subsection{Model fitting}

Function \code{hglm} is the main function for fitting HGLMs within package 
\pkg{holiglm}.
\begin{verbatim}
hglm(formula, family = gaussian(), data, constraints = NULL, 
  weights = NULL, scaler = c("auto", "center_standardization", 
    "center_minmax", "standardization", "minmax", "off"), 
  scale_response = NULL, big_m = 100, solver = "auto", control = list(), 
  dry_run = FALSE) 
\end{verbatim}
The design of the function arguments resembles that of the \code{stats::glm()} function,
with some additional arguments.
\begin{description}
  \item[\code{constraints}] The constraints imposed on the GLM. All constraints 
    are of class \code{"hglmc"}, so the \code{constraints} argument expects
    a list of \code{"hglmc"} constraints.
  \item[\code{scaler}] Especially for constraints which rely on the big-$M$ 
    formulation, scaling is very important as it helps to make reliable choices 
    in regard to the chosen big-$M$ value.
  \item[\code{scale\_response}] Whether the response shall be standardized.
    Scaling the response is only possible for family \code{gaussian()}, where the default
    is also to scale it.
  \item[\code{big\_m}] An upper bound $M$ for the coefficients, needed for the 
    big-$M$ constraint.
  \item[\code{solver}] By default, the best available optimization solver is chosen
    automatically. However, it is possible to force the use of a particular solver
    by providing the solver name. The argument is then passed along to \code{ROI::ROI_solve}.
  \item[\code{control}] The control argument can be used to pass additional
    arguments along to \code{ROI::ROI_solve}.
  \item[\code{dry\_run}] The \code{dry\_run} argument allows to obtain the underlying
    optimization problem. This can be useful if the user wants to impose additional
    constraints, which are currently not implemented, directly to the \pkg{ROI}
    optimization object.
\end{description}

By making use of the formula system provided by the \pkg{stats} package, we ensure that
all the typical operations work as in \code{stats::glm}.
We illustrate the usage of \code{hglm} without constraints for the log-binomial
regression model. This particular family-link combination was chosen as a first
example as it can be shown (by simulation) that using conic programming instead of the 
iteratively reweighted least squares (IRLS) employed by \code{glm} is not only more reliable
but potentially faster \citep{log-bin:schwendinger:2021}.
\begin{Schunk}
\begin{Sinput}
R> data("Caesarian", package = "lbreg")
R> fit <- hglm(cbind(n1, n0) ~ RISK + NPLAN + ANTIB, 
+    binomial(link = "log"), solver = "ecos", 
+    data = Caesarian, constraints = NULL)
R> fit
\end{Sinput}
\begin{Soutput}

Call:  hglm(formula = cbind(n1, n0) ~ RISK + NPLAN + ANTIB, 
    binomial(link = "log"), data = Caesarian, constraints = NULL,
    solver = "ecos")

Coefficients:
(Intercept)         RISK        NPLAN        ANTIB  
     -1.977        1.295        0.545       -2.066  

Degrees of Freedom: 6 Total (i.e. Null);  3 Residual
Null Deviance:	    83.49 
Residual Deviance: 6.308 	AIC: 31.49
\end{Soutput}
\end{Schunk}
Given that the \code{"hglm"} object inherits from class \code{"glm"}, the output 
of the \code{summary} function is similar to the one from \code{stats::glm}: 
\begin{Schunk}
\begin{Sinput}
R> summary(fit)
\end{Sinput}
\begin{Soutput}

Call:
hglm(formula = cbind(n1, n0) ~ RISK + NPLAN + ANTIB, 
    binomial(link = "log"), data = Caesarian, constraints = NULL,
    solver = "ecos")

Deviance Residuals: 
       1         2         3         4         5         6         8  
 1.06853  -0.34543  -2.21588   0.20040  -0.26605  -0.15022   0.05687  

Coefficients:
            Estimate Std. Error z value Pr(>|z|)    
(Intercept)  -1.9774     0.3311  -5.972 2.35e-09 ***
RISK          1.2950     0.3407   3.801 0.000144 ***
NPLAN         0.5450     0.1451   3.755 0.000174 ***
ANTIB        -2.0659     0.2814  -7.342 2.11e-13 ***
---
Signif. codes:  0 '***' 0.001 '**' 0.01 '*' 0.05 '.' 0.1 ' ' 1

(Dispersion parameter for binomial family taken to be 1)

    Null deviance: 83.4914  on 6  degrees of freedom
Residual deviance:  6.3079  on 3  degrees of freedom
AIC: 31.489

Number of iterations: 26
\end{Soutput}
\end{Schunk}
In the above, the number of iterations is the one needed by the solver to reach convergence.

Instead of straight using function \code{hglm} for estimating the model, 
it is also possible to take a different approach, especially if the modeler wants 
to access the underlying optimization problem directly:
\begin{enumerate}
  \item Generate the model with \code{hglm_model}.
  \item Generate the underlying optimization problem with \code{as.OP}.
  \item Fit the model with \code{hglm_fit}.
\end{enumerate}
To generate the model,
the function \code{hglm_model} can be used:
\begin{Schunk}
\begin{Sinput}
R> x <- model.matrix(cbind(n1, n0) ~ RISK + NPLAN + ANTIB, data = Caesarian)
R> model <- hglm_model(x = x, y = with(Caesarian, cbind(n1, n0)), 
+                      binomial(link = "log"))
\end{Sinput}
\end{Schunk}
Alternatively, \code{hglm} with the argument \code{dry\_run} set to \code{TRUE}
can be used to obtain the optimization model:
\begin{Schunk}
\begin{Sinput}
R> model <- hglm(cbind(n1, n0) ~ RISK + NPLAN + ANTIB, data = Caesarian, 
+                binomial(link = "log"), constraints = NULL, dry_run = TRUE)
\end{Sinput}
\end{Schunk}
After the model is set up, the underlying optimization problem can be built
by using the generic function \code{as.OP} exported from \pkg{ROI} and solved
using function \code{ROI\_solve} in \pkg{ROI}.
\begin{Schunk}
\begin{Sinput}
R> op <- as.OP(model)
R> print(op)
\end{Sinput}
\begin{Soutput}
ROI Optimization Problem:

Minimize a linear objective function of length 18 with
- 18 continuous objective variables,

subject to
- 49 constraints of type conic.
  |- 42 conic constraints of type 'expp'
  |- 7 conic constraints of type 'nonneg'
- 18 lower and 0 upper non-standard variable bounds.
\end{Soutput}
\begin{Sinput}
R> ROI::ROI_solve(op)
\end{Sinput}
\begin{Soutput}
Optimal solution found.
The objective value is: 1.109144e+02
\end{Soutput}
\end{Schunk}
Alternatively, the model can be fitted with \code{hglm\_fit}. Here the 
solver output is already converted into a form similar to the output of 
\code{glm.fit}.
\begin{Schunk}
\begin{Sinput}
R> fit <- hglm_fit(model)
R> fit
\end{Sinput}
\begin{Soutput}

Call:  NULL

Coefficients:
(Intercept)         RISK        NPLAN        ANTIB  
     -1.977        1.295        0.545       -2.066  

Degrees of Freedom: 6 Total (i.e. Null);  3 Residual
Null Deviance:	    83.49 
Residual Deviance: 6.308 	AIC: 31.49
\end{Soutput}
\end{Schunk}
%%%%%%%%%%%%%%%%%%%%%%%%%%%%%%%%%%%%%%%%%%%%%%%%%%%%%%%%%%%%%%%%%%%%%%%%%%%%%%%%%%%%%%%%%%%%%%%%%%%%%%%%%%%%%%%%%%%%%%%%%%%%%%%%%%%%%%%%%%%%
% Constraints
%%%%%%%%%%%%%%%%%%%%%%%%%%%%%%%%%%%%%%%%%%%%%%%%%%%%%%%%%%%%%%%%%%%%%%%%%%%%%%%%%%%%%%%%%%%%%%%%%%%%%%%%%%%%%%%%%%%%%%%%%%%%%%%%%%%%%%%%%%%%
\subsection{Constraints}
The most distinctive feature of the \pkg{holiglm} package is that it allows for
the usage of many predefined constraints. 
In Section~\ref{sec:holistic_constraints} the constraints were introduced. 
In this section we focus on the \proglang{R} usage of the constraints.
Table~\ref{tab:overview_constraints} gives 
an overview of all the available constraints in \pkg{holiglm}.
%\begin{landscape}
\begin{table}[t!]
\centering
  \begin{tabular}{@{}llp{5.5cm}@{}}
   \hline
    Constraint &
        Category &
            Description \\
    \hline
%    \ \\
    \code{k_max(k)}$^a$ &
        \hyperref[con:global_sparsity]{global sparsity} &
            Upper bound on the number of active covariates.
        \\
    \code{group_sparsity(vars, k = 1L)}$^a$ &
        \hyperref[con:group_sparsity]{group sparsity} &
            Upper bound on the number of active covariates within a group of 
            covariates  \code{vars}.
        \\
    \code{rho_max(rho)}$^a$ &
        \hyperref[con:rho_max]{group sparsity} &
            Upper bound on the strength of the pairwise correlation;
            for pairs which exceed this bound only one of the two variables 
            is allowed in the model.
        \\
    \code{group_inout(vars)}$^a$ &
        \hyperref[con:group_sparsity]{group sparsity} &
            Forces all covariates within a group are either in (all coefficients are non-zero) 
            or out (all coefficients are zero) of the model.
        \\
    \code{include(vars)}$^a$ &
        \hyperref[con:modeler_expertise]{modeler expertise} &
            Forces the inclusion of covariates.
        \\
    \code{lower(kvars)}$^b$  &
        \hyperref[con:bounded_domains]{bounded domains} &
            Forces lower bounds on coefficients.
        \\
    \code{upper(kvars)}$^b$  &
        \hyperref[con:bounded_domains]{bounded domains} &
            Forces upper bounds on coefficients.
        \\
    \code{linear(L, dir, rhs)}$^b$  &
        \hyperref[con:linear_constraint]{linear constraint} &
            Imposes linear constraints on the coefficients or on the $z$ variables
            from the big-$M$ constraint.
        \\
    \code{group_equal(vars)}$^b$  &
        \hyperref[con:linear_constraint]{linear constraint} &
            Forces all covariates in the specified group to have the same coefficient.
        \\
    \code{sign_coherence(vars)}$^a$ &
        \hyperref[con:sign_coherence]{sign coherence} &
            Ensures that the coefficients of the specified covariates have
            the same sign.
        \\
    \code{pairwise_sign_coherence(rho)}$^a$ &
        \hyperref[con:sign_coherence]{sign coherence} &
            Ensures that coefficients of covariates that exhibit pairwise 
            correlation more than \code{rho} (in absolute value) have the same 
            sign.
        \\
       \hline
  \end{tabular}
  \caption[Available constraints]{Overview on the pre-implemented constraints. Here only
    the main arguments are shown, all arguments can be found in the \pkg{holiglm} manual.
    The constraints marked with $^a$ are of mixed-integer type, while the ones marked with 
    $^b$ are of linear type.}
  \label{tab:overview_constraints}
\end{table}
%\end{landscape}
\subsubsection{Combining constraints}
Constraints can be combined by the combine operator \code{c()}. 
In the following example, we ensure that at most $5$ covariates are in the final 
model and that the pairwise correlation of the active variables is less than or 
equal to $0.8$ by combining the two constraints:
\begin{Schunk}
\begin{Sinput}
R> c(k_max(5), rho_max(0.8))
\end{Sinput}
\begin{Soutput}
List of Holistic GLM Sparsity Constraints
[[1]]
Holistic GLM Sparsity Constraint of Type 'k_max'
List of 1
 $ k_max: int 5

[[2]]
Holistic GLM Sparsity Constraint of Type 'rho_max'
List of 4
 $ rho_max: num 0.8
 $ exclude: chr "(Intercept)"
 $ use    : chr "everything"
 $ method : chr "pearson"
\end{Soutput}
\end{Schunk}
\subsubsection{Global sparsity}
The global sparsity constraint \code{k_max} can be used to enforce an upper bound
on the number of non-zero covariates contained in the model. The upper bound is
enforced on the number of columns in the model matrix without taking into account the intercept.
\begin{Schunk}
\begin{Sinput}
R> fit <- hglm(cbind(n1, n0) ~ RISK + NPLAN + ANTIB, binomial(link = "log"),
+    constraints = k_max(2), data = Caesarian)
R> coef(fit)
\end{Sinput}
\begin{Soutput}
(Intercept)        RISK       NPLAN       ANTIB 
  -1.818324    1.319768    0.000000   -1.774368 
\end{Soutput}
\end{Schunk}
As one can see by looking at the coefficients, in contrast to other sparsity-inducing methods,
the coefficients are not just close to zero but exactly 
zero if not included in the model. The information about the active 
coefficients can be obtained by using the \code{active_coefficients} method.
\begin{Schunk}
\begin{Sinput}
R> active_coefficients(fit)
\end{Sinput}
\begin{Soutput}
(Intercept)        RISK       NPLAN       ANTIB 
       TRUE        TRUE       FALSE        TRUE 
\end{Soutput}
\end{Schunk}
\subsubsection{Group sparsity}
The group sparsity constraint can be used to restrict the number of covariates
to be included from a particular subgroup. The \code{vars} argument is a vector 
containing the names of the variables for which the constraint should be applied. 
Note that the names of the variables should correspond to the column names of 
the model matrix. As previously mentioned, a possible use case for this type of 
constraint is the selection of the best out of different transformations. 
For illustration, we employ a Poisson regression with log link to model the 
number of apprentices moving to Edinburgh from different counties. 
Data is available in package \pkg{GLMsData} \citep{pkg:GLMsData:2018}. 
\begin{Schunk}
\begin{Sinput}
R> data("apprentice", package = "GLMsData")
\end{Sinput}
\end{Schunk}
We specify the following model with group constraints. For the variables
distance \code{Dist} and population \code{Pop} also the log transformations are considered for inclusion in the model.
\begin{Schunk}
\begin{Sinput}
R> fo <- Apps ~ Dist + log(Dist) + Pop + log(Pop) + Urban + Locn
R> constraints <- c(group_sparsity(c("Dist", "log(Dist)")), 
+    group_sparsity(c("Pop", "log(Pop)")))
R> fit <- hglm(fo, constraints = constraints, family=poisson(),
+    data = apprentice)
R> coef(fit)
\end{Sinput}
\begin{Soutput}
(Intercept)        Dist   log(Dist)         Pop    log(Pop) 
 7.51921885  0.00000000 -2.05542948  0.00000000  0.98217367 
      Urban   LocnSouth    LocnWest 
-0.01705048  0.50439145 -0.10157628 
\end{Soutput}
\end{Schunk}
Indeed, the logarithmically transformed variables are selected for inclusion in
the model.

In-out constraints implemented in \code{group\_inout} force the coefficients 
of all variables in argument \code{vars} to be jointly zero or different than zero. An application
for this type of constraints is dummy-encoded categorical features, where 
either all dummy variables should be included in the model or none.
The following Poisson regression model is restricted to include three covariates, 
whereas all dummies 
corresponding to the variable \code{Locn} (location relative to Edinburgh)
should be included or excluded from the model:
\begin{Schunk}
\begin{Sinput}
R> fo <- Apps ~ log(Dist) + log(Pop) + Urban + Locn
R> constraints <- c(k_max(3), group_inout(c("LocnSouth", "LocnWest")))
R> fit <- hglm(fo, constraints = constraints, 
+    family = poisson(), data = apprentice)
\end{Sinput}
\end{Schunk}
To restricting the pairwise collinearity,
we provide the constraint \code{rho_max} which sets an upper bound on the
pairwise collinearity. 
For the \code{Boston} data set from the \pkg{MASS} package \citep{book:mass:2002} 
we specify the following linear model, where the empirical correlation of 
variables \code{tax} and \code{rad} is more than $0.9$:
\begin{Schunk}
\begin{Sinput}
R> data("Boston", package = "MASS")
R> fit <- hglm(medv ~ rm + rad + dis + lstat + tax + ptratio, 
+    data = Boston, constraints = rho_max(0.9))
R> coef(fit)
\end{Sinput}
\begin{Soutput}
 (Intercept)           rm          rad          dis        lstat 
23.790456303  4.323450402  0.000000000 -0.695741701 -0.623392021 
         tax      ptratio 
-0.005568159 -0.846772468 
\end{Soutput}
\end{Schunk}
The variable \code{rad} is excluded from the model.

\subsubsection{Modeler expertise}
The function \code{include} can be used to force the inclusion of a specific
variable in the model. This can be useful if a modeler is interested in setting 
an upper bound on the number of active variables while ensuring that a certain variable stays in the model. For example, the following model on the 
\code{Caesarian} data set will include two variables and one of them will be \code{NPLAN}:
\begin{Schunk}
\begin{Sinput}
R> fit <- hglm(cbind(n1, n0) ~ RISK + NPLAN + ANTIB, binomial(link = "log"),
+    constraints = c(k_max(2), include("NPLAN")), data = Caesarian)
R> coef(fit)
\end{Sinput}
\begin{Soutput}
(Intercept)        RISK       NPLAN       ANTIB 
 -1.0056143   0.0000000   0.5892485  -1.7899819 
\end{Soutput}
\end{Schunk}
\subsubsection{Bounded domains}
The functions \code{lower} and \code{upper} can be used to set lower and upper
bounds on the value of the coefficients. This can be used, for example, to add 
non-negativity constraints or any other bounds on the coefficients.
\begin{Schunk}
\begin{Sinput}
R> constraints <- c(lower(c(RISK = 3, ANTIB = 1e-3)), upper(c(NPLAN = -1)))
R> fit <- hglm(cbind(n1, n0) ~ RISK + NPLAN + ANTIB, binomial(link = "log"),
+    constraints = constraints, data = Caesarian)
R> coef(fit)
\end{Sinput}
\begin{Soutput}
(Intercept)        RISK       NPLAN       ANTIB 
   -3.67126     3.00000    -1.00000     0.00100 
\end{Soutput}
\end{Schunk}
Note that all three constraints are binding.

\subsubsection{Linear constraints}
To impose linear constraints on the coefficients, the function \code{linear} can 
be used. In the following example we will enforce that 
$\beta_\text{RISK} \leq \beta_\text{ANTIB}$ in the log-binomial regression model:
\begin{Schunk}
\begin{Sinput}
R> risk_leq_antib <- c(RISK = 1, ANTIB = -1)
R> fit <- hglm(cbind(n1, n0) ~ RISK + NPLAN + ANTIB, binomial(link = "log"),
+    constraints = linear(risk_leq_antib, "<=", 0), data = Caesarian)
R> coef(fit)
\end{Sinput}
\begin{Soutput}
(Intercept)        RISK       NPLAN       ANTIB 
-0.95854528 -0.30446220  0.09905912 -0.30446220 
\end{Soutput}
\end{Schunk}
The next example shows how to enforce that the coefficients of \code{RISK} and
\code{NPLAN} sum to one, i.e. $\beta_\text{RISK} + \beta_\text{NPLAN} = 1$.
\begin{Schunk}
\begin{Sinput}
R> risk_nplan <- c(RISK = 1, NPLAN = 1)
R> fit <- hglm(cbind(n1, n0) ~ RISK + NPLAN + ANTIB, binomial(link = "log"),
+    constraints = linear(risk_nplan, "==", 1), data = Caesarian)
R> coef(fit)
\end{Sinput}
\begin{Soutput}
(Intercept)        RISK       NPLAN       ANTIB 
 -1.3047691   0.6600015   0.3399985  -1.9232422 
\end{Soutput}
\begin{Sinput}
R> coef(fit)[["RISK"]] + coef(fit)[["NPLAN"]]
\end{Sinput}
\begin{Soutput}
[1] 1
\end{Soutput}
\end{Schunk}
To impose both constraints at the same time, we can either combine the linear 
constraints or use the matrix notation as shown below.
\begin{Schunk}
\begin{Sinput}
R> L <- rbind(c(RISK = 1, NPLAN = 0, ANTIB = -1), c(1, 1, 0))
R> fit <- hglm(cbind(n1, n0) ~ RISK + NPLAN + ANTIB, binomial(link = "log"),
+    constraints = linear(L, c("<=", "=="), c(0, 1)), data = Caesarian)
R> coef(fit)
\end{Sinput}
\begin{Soutput}
(Intercept)        RISK       NPLAN       ANTIB 
 -1.8274949  -0.2560339   1.2560339  -0.2560339 
\end{Soutput}
\end{Schunk}
A special case of linear constraint is \code{group\_equal}, which ensures that
the coefficients of the variables in \code{vars} are equal:
\begin{Schunk}
\begin{Sinput}
R> fit <- hglm(cbind(n1, n0) ~ RISK + NPLAN + ANTIB, binomial(link = "log"),
+    constraints = group_equal(c("RISK", "NPLAN", "ANTIB")), 
+    data = Caesarian)
R> coef(fit)
\end{Sinput}
\begin{Soutput}
(Intercept)        RISK       NPLAN       ANTIB 
 -0.9710750  -0.1750262  -0.1750262  -0.1750262 
\end{Soutput}
\end{Schunk}

Finally, we showcase how linear constraints can be imposed on the $z$ variables by
setting \code{on_big_m = TRUE} in function \code{linear}.
This can be useful for modeling restrictions of the type 
``if variable $j$ is included variable $k$ should also be included''.
An application for this type of constraints is estimating models with 
polynomial features,  where if a higher order of the polynomial is included 
in the model, it is desirable that all lower orders are also included. 
The following linear model example uses data \code{heatcap} from package 
\pkg{GLMsData} where the relation of heat capacity for
a type of acid and temperature is investigated. 
\begin{Schunk}
\begin{Sinput}
R> data("heatcap", package = "GLMsData")
\end{Sinput}
\end{Schunk}
We consider polynomial features up to degree five:
\begin{Schunk}
\begin{Sinput}
R> fo <- Cp ~ poly(Temp, 5)
\end{Sinput}
\end{Schunk}
The linear constraint below corresponds to the constraint
$$
z_{\text{\code{Temp}}^5}
\leq z_{\text{\code{Temp}}^4}
\leq z_{\text{\code{Temp}}^3}
\leq z_{\text{\code{Temp}}^2}
\leq z_{\text{\code{Temp}}},
$$
which ensures that a lower order polynomial feature is included if all
preceding (higher order) ones are included in the model.
\begin{Schunk}
\begin{Sinput}
R> L <- rbind(c(-1, 1, 0, 0, 0), c(0, -1, 1, 0, 0), 
+    c(0, 0, -1, 1, 0), c(0, 0, 0, -1, 1))
R> colnames(L) <- colnames(model.matrix(fo, data = heatcap))[-1]
R> lin_z_c <- linear(L, rep("<=", nrow(L)), rep(0, nrow(L)), on_big_m = TRUE)
\end{Sinput}
\end{Schunk}
When additionally setting $k_\text{max}=3$, we see that all polynomial features 
up to degree $3$ will be included in the model.
\begin{Schunk}
\begin{Sinput}
R> constraints <- c(k_max(3), lin_z_c)
R> fit <- hglm(fo, data = heatcap, constraints = constraints)
R> coef(fit)
\end{Sinput}
\begin{Soutput}
   (Intercept) poly(Temp, 5)1 poly(Temp, 5)2 poly(Temp, 5)3 
    11.2755556      1.8409086      0.3968900      0.1405174 
poly(Temp, 5)4 poly(Temp, 5)5 
     0.0000000      0.0000000 
\end{Soutput}
\end{Schunk}
\subsubsection{Sign coherence constraint}
We estimate a Poisson model with identity link for the \code{apprentice} data where 
we use interactions of the location with the other covariates. In this case
it can be desirable that, for each covariate, the slopes for all locations have 
the same sign. Note that the model includes no intercept:
\begin{Schunk}
\begin{Sinput}
R> fo <- Apps ~  0 + Locn + Locn : Dist +  Locn : Pop + Locn : Urban
R> constraints <- c(
+    sign_coherence(c("LocnNorth:Dist", "LocnSouth:Dist", "LocnWest:Dist")),
+    sign_coherence(c("LocnNorth:Pop", "LocnSouth:Pop", "LocnWest:Pop")),
+    sign_coherence(c("LocnNorth:Urban", "LocnSouth:Urban", "LocnWest:Urban")))
R> fit <- hglm(fo, constraints = constraints, family = poisson("identity"), 
+    data = apprentice)
R> coef(fit)
\end{Sinput}
\begin{Soutput}
      LocnNorth       LocnSouth        LocnWest  LocnNorth:Dist 
   4.929168e-02   -1.325961e+01    1.159558e+01   -6.226539e-03 
 LocnSouth:Dist   LocnWest:Dist   LocnNorth:Pop   LocnSouth:Pop 
  -2.172583e-01   -9.535403e-02    1.458556e-01    1.618307e+00 
   LocnWest:Pop LocnNorth:Urban LocnSouth:Urban  LocnWest:Urban 
   8.259307e-02    9.274395e-03    5.071724e-01    8.709769e-07 
\end{Soutput}
\end{Schunk}
The slope parameters for distance are all negative, while
the slope parameters for population and degree of urbanization are all positive. 

Finally, we showcase on the \code{Boston} data set how we can restrict the highly
correlated covariates to have the same sign using \code{pairwise\_sign\_coherence}:
\begin{Schunk}
\begin{Sinput}
R> data("Boston", package = "MASS")
R> fit <- hglm(medv ~ rm + rad + dis + lstat + tax + ptratio, 
+    data = Boston, constraints = pairwise_sign_coherence(0.8))
R> coef(fit)
\end{Sinput}
\begin{Soutput}
  (Intercept)            rm           rad           dis         lstat 
 2.379045e+01  4.323451e+00 -3.998741e-07 -6.957417e-01 -6.233920e-01 
          tax       ptratio 
-5.568140e-03 -8.467723e-01 
\end{Soutput}
\end{Schunk}
Unlike in the unrestricted model, variables \code{rad} and \code{tax} both have
a negative sign. Note also that, in this example, the coefficient of \code{rad}
is close to zero, which is in line with the result obtained when using 
\code{rho\_max} constraint.

\subsection{Auxiliary functions}

\subsubsection{Estimate a sequence of models}
Function \code{hglm_seq} can be used to estimate the sequence of models
containing $k=1, \ldots, p$ covariates. The print method for the resulting \code{"hglm_seq"} object includes the AIC, BIC as well as the vector of 
coefficients for each model. For the polynomial regression example introduced 
before, we can estimate the sequence of models:
\begin{Schunk}
\begin{Sinput}
R> fit_seq <- hglm_seq(formula = Cp ~ poly(Temp, 5), data = heatcap,
+    constraints = lin_z_c)
R> fit_seq
\end{Sinput}
\begin{Soutput}
HGLM Fit Sequence:
  k_max    aic    bic (Intercept) poly(Temp, 5)1 poly(Temp, 5)2
1     5 -64.52 -58.29    11.27556       1.840909        0.39689
2     4 -65.60 -60.26    11.27556       1.840909        0.39689
3     3 -64.04 -59.59    11.27556       1.840909        0.39689
4     2 -52.30 -48.74    11.27556       1.840909        0.39689
5     1 -24.42 -21.75    11.27556       1.840909        0.00000
  poly(Temp, 5)3 poly(Temp, 5)4 poly(Temp, 5)5
1      0.1405174    -0.05560884     0.02653116
2      0.1405174    -0.05560884     0.00000000
3      0.1405174     0.00000000     0.00000000
4      0.0000000     0.00000000     0.00000000
5      0.0000000     0.00000000     0.00000000
\end{Soutput}
\end{Schunk}
We can observe that the model with polynomial features up to degree $4$ is
the one preferred by both AIC and BIC.

\subsubsection{Group duplicates}

For binomial regression models, if all the covariates are 
categorical or discrete variables with few possible values, 
it is often the case that the model matrix contains duplicated rows. 
In this case, it is convenient to aggregate the original data to a
data frame containing the unique values of the covariates in the rows and for
each row the number of successes and failures corresponding to the response. 
Function \code{agg_binomial} can be used for this purpose
to reduce the number of rows from originally 16949 to 
74:
\begin{Schunk}
\begin{Sinput}
R> data("Heart", package = "lbreg")
R> heart <- agg_binomial(Heart ~ ., data = Heart, as_list = FALSE)
R> c(nrow(Heart), nrow(heart))
\end{Sinput}
\begin{Soutput}
[1] 16949    74
\end{Soutput}
\end{Schunk}
This function should always be used when possible, as it 
drastically reduces the size of the underlying
optimization problem and therefore, reduces the solving time. The new function 
call would be:
\begin{Schunk}
\begin{Sinput}
R> hglm(cbind(success, failure) ~ ., binomial(link = "log"),
+    data = heart, constraints = NULL)
\end{Sinput}
\end{Schunk}
%
% ------------------------------------------------------------------------------
%
% Use cases
%
% ------------------------------------------------------------------------------
\section{Use cases}\label{sec:use_cases}
This section illustrates the usage of the \pkg{holiglm} package on 
more realistic real-world examples.

\subsection{Fairness in logistic regression}
We illustrate how the \pkg{holiglm} package can be employed for 
estimating the coefficients of a logistic regression model with the fairness 
constraints proposed in \cite{zafar_fairness_2019}. 
In the fairness literature, one is typically interested in building classification
models free of the so-called \emph{disparate impact}.
A decision-making process suffers from disparate impact if it grants
disproportionately large fraction of beneficial (or positive) 
outcomes to certain groups of a sensitive feature, such as gender or race. 
Assuming we have a binary response variable $y$ and a binary sensitive feature 
$w$, a binary classifier is free of disparate impact if the probability that the 
classifier assigns an observation to the positive class is the same for both 
values of the sensitive feature $w$: $\Prob(\hat y=1|w=0)=\Prob(\hat y=1|w=1)$. 
If $w$ has more than two classes, this relation should hold for all values of $w$. 
Note that even if the sensitive variable $w$ is not included in the model, the 
classifier can still suffer from disparate impact (mainly due to the association
between the non-sensitive variables $\boldsymbol x$ included in the model and the 
sensitive variable).

However, for many classifiers these probabilities are typically nonconvex 
functions of the parameters, which complicates estimation.
To provide a more general framework, \cite{zafar_fairness_2019} propose a 
covariance measure of decision boundary unfairness. This is given by the 
covariance between a sensitive binary attribute $w$ and the signed distance 
from the (non-sensitive) feature vectors $\boldsymbol x_i$ to the decision boundary:
\begin{equation}\label{eq:fair:emp_cov}
\COV(w, d_{\boldsymbol\beta}(\boldsymbol x))=
 \E((w-\bar w) d_{\boldsymbol\beta}(\boldsymbol x))-
 \underbrace{\E((w-\bar w)) \E(d_{\boldsymbol\beta}(\boldsymbol x))}_{=0}
 \approx \frac{1}{n}\sum_{i=1}^n(w_i-\bar w)d_{\boldsymbol\beta}
                                              (\boldsymbol x_i).
\end{equation}
The framework can also accommodate for various binary sensitive variables $w^{(k)}$, 
$k=1,\ldots, K$, where limits are imposed for the above covariance measure for each
of the $k$ sensitive variables. For the logistic regression problem, where the signed distance 
to the decision boundary is proportional to $\boldsymbol x_i^\top\boldsymbol\beta$, 
the fairness constrained model is given by:
\begin{equation}\label{eq:fair}
\begin{array}{rl}
\underset{\boldsymbol\beta}{\text{maximize}}  &  \sum_{i=1}^n \log(p(y_i|\boldsymbol x_i,
  \boldsymbol\beta))  \\
\text{subject to}
    & \frac{1}{n}\sum_{i=1}^n(w^{(1)}_i-\bar w^{(1)}) 
          \boldsymbol x_i^\top\boldsymbol\beta \leq c_1 \\
    & \frac{1}{n}\sum_{i=1}^n(w^{(1)}_i-\bar w^{(1)})
          \boldsymbol x_i^\top\boldsymbol\beta  \geq -c_1,\\
    & ... \\
    & \frac{1}{n}\sum_{i=1}^n(w^{(K)}_i-\bar w^{(K)}) 
          \boldsymbol x_i^\top\boldsymbol\beta \leq c_K \\
    & \frac{1}{n}\sum_{i=1}^n(w^{(K)}_i-\bar w^{(K)}) 
          \boldsymbol x_i^\top\boldsymbol\beta \geq -c_K,
\end{array}
\end{equation}
where $c_k\in \mathbb{R}^+$ for $k = 1, \ldots, K$  is a given constant which 
trades off accuracy and decision boundary unfairness.

As an illustrative example we employ the \code{AdultUCI} data set in package 
\pkg{arules} \citep{arules}, which contains data extracted from the 1994 U.S. 
census bureau database. The response variable in this data set is the binary 
variable \code{income} which indicates whether the income of the respondents 
exceeds USD $50\,000$ per year. 
\begin{Schunk}
\begin{Sinput}
R> data("AdultUCI", package = "arules")
\end{Sinput}
\end{Schunk}
Before estimation, we eliminate all observations with missing values:
\begin{Schunk}
\begin{Sinput}
R> AdultUCI <- na.omit(AdultUCI)
R> dim(AdultUCI)
\end{Sinput}
\begin{Soutput}
[1] 30162    15
\end{Soutput}
\end{Schunk}
The data is rather unbalanced in terms of the distribution of the response:
\begin{Schunk}
\begin{Sinput}
R> prop.table(table(AdultUCI$income))
\end{Sinput}
\begin{Soutput}

    small     large 
0.7510775 0.2489225 
\end{Soutput}
\end{Schunk}
Moreover, this data set is known to be skewed in terms of \code{race}
and \code{gender}. 
\begin{Schunk}
\begin{Sinput}
R> xtabs(~ sex + race, data = AdultUCI)
\end{Sinput}
\begin{Soutput}
        race
sex      Amer-Indian-Eskimo Asian-Pac-Islander Black Other White
  Female                107                294  1399    87  7895
  Male                  179                601  1418   144 18038
\end{Soutput}
\end{Schunk}
Given the sparsely populated \code{race} groups \code{Amer-Indian-Eskimo}, 
\code{Asian-Pac-Islander} and \code{Other}, we merge them into the \code{Other} 
group.
\begin{Schunk}
\begin{Sinput}
R> levels(AdultUCI$race)[c(1, 2)] <- "Other"
\end{Sinput}
\end{Schunk}
Clear differences are observable in the percentage of high income 
earners in the \code{sex} and \code{race} groups:
\begin{Schunk}
\begin{Sinput}
R> aggregate(income ~ sex + race, data = AdultUCI, 
+    function(x) mean(as.numeric(x) - 1))
\end{Sinput}
\begin{Soutput}
     sex  race     income
1 Female Other 0.11475410
2   Male Other 0.26731602
3 Female Black 0.06075768
4   Male Black 0.19816643
5 Female White 0.12298923
6   Male White 0.32531323
\end{Soutput}
\end{Schunk}
We will therefore consider both \code{race} and \code{gender} as sensitive
variables in our use case. We create one $w^{(k)}$ binary (i.e., dummy) 
variable for each factor level combination.
\begin{Schunk}
\begin{Sinput}
R> W <- model.matrix(~ 0 + sex:race, data = AdultUCI)
\end{Sinput}
\end{Schunk}
We consider the following predictive model for the analysis (note that the
sensitive variables are not included in the model formula):
\begin{Schunk}
\begin{Sinput}
R> form <- "income ~ age + relationship + `marital-status` + workclass +
+    `education-num` + occupation + I(`capital-gain` - `capital-loss`) +
+    `hours-per-week`"
\end{Sinput}
\end{Schunk}
The variable \code{native-country} is not included in the model as it was 
eliminated in a preliminary stepwise selection procedure.

\begin{figure}[t!]
\centering
\begin{Schunk}

\includegraphics[width=\maxwidth]{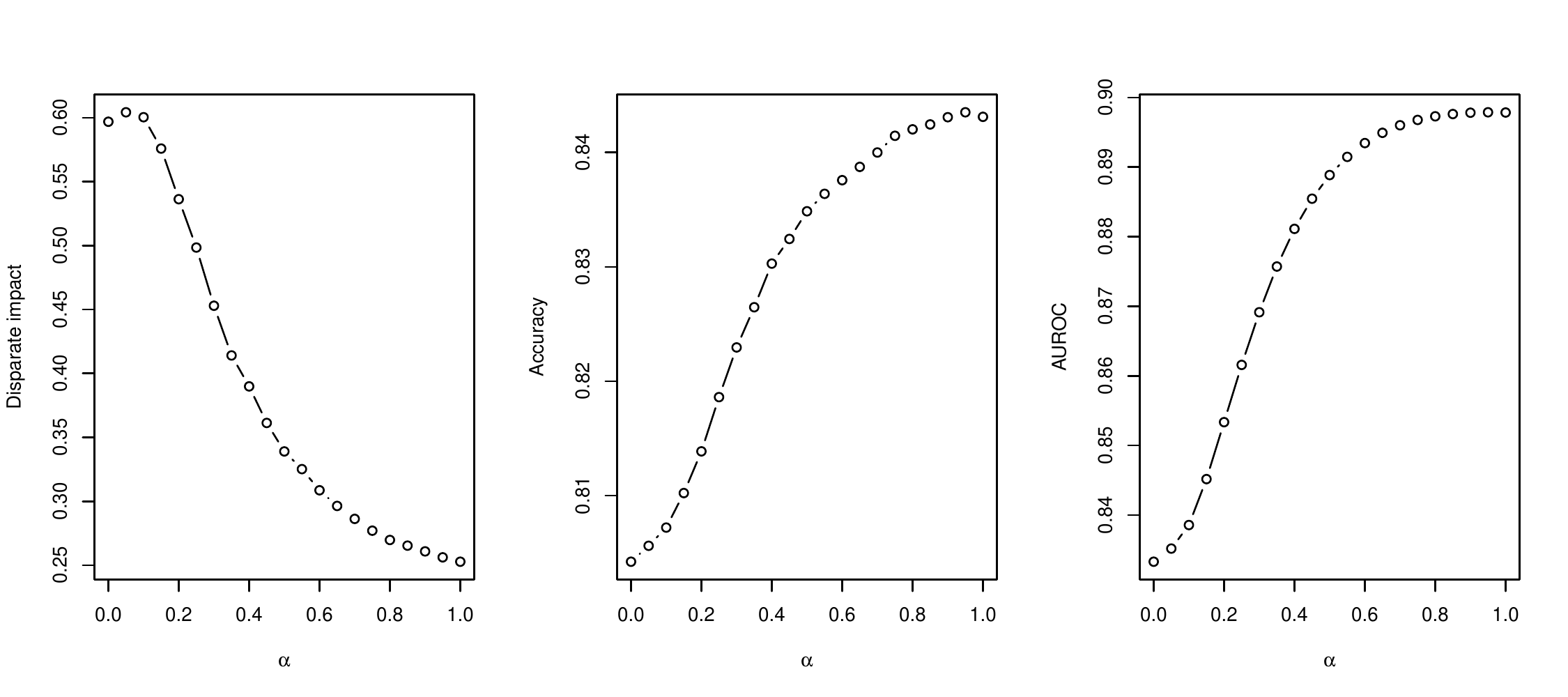} \end{Schunk}
\caption{\label{fig:fair:comparison} This figure displays the disparate impact,
accuracy and area under the ROC curve for different degrees of fairness 
given by $\alpha\in [0,1],$ where an $\alpha$ of zero implies a fair model in 
terms of the specified covariance measure.}
\end{figure}

In order to assess the trade-off between fairness and prediction accuracy, we 
follow \cite{zafar_fairness_2019} and choose a grid of values 
$\alpha_k \in [0,1]$ which are used to compute $c_k=\alpha_k s^{(k)}$. Here 
$s^{(k)}$ is the empirical covariance (in absolute value) between the sensitive 
variable $w^{(k)}$ and the linear predictor $\boldsymbol x_i^\top \hat{\boldsymbol\beta}$ 
from the unconstrained regression; $s^{(k)}$ serves in this case as an upper 
bound on the disparate impact covariance measure.
A value $\alpha_k=0$ represents the case where the empirical covariance in
Equation~\ref{eq:fair:emp_cov} is restricted to zero while $\alpha_k=1$ 
represents the case with no fairness constraints. We first estimate the 
unconstrained model using the  \code{stats::glm} function and the calculate 
$s^{(k)}$: 
\begin{Schunk}
\begin{Sinput}
R> m0 <- glm(as.formula(form), family = binomial(), data = AdultUCI)
R> s <- apply(W, 2, function(w) abs(cov(w, m0$linear.predictors)))
\end{Sinput}
\end{Schunk}
The left-hand side of the constraints from Equation~\ref{eq:fair} can be 
specified in matrix form $L\boldsymbol\beta$ where $L$ is constructed by:
\begin{Schunk}
\begin{Sinput}
R> xm <- model.matrix(m0)
R> L <- t(apply(W, 2, function(w) colMeans((w - mean(w)) * xm)))
\end{Sinput}
\end{Schunk}
For a grid of $\alpha_k$ values, we estimate the fairness constrained logistic
regressions. We set the \code{big_m} argument to $5$, which means that the 
standardized coefficients (see \code{scaler = "standardization"}) are constrained
to be at most $5$ in absolute value. A value of $5$ for \code{big_m} is small 
enough to ensure a fast convergence of the algorithm and, at the same time,
sufficiently large to be non-restrictive for the problem at hand
(in none of the estimated models this bound was reached). 
For each model we store the in-sample predictions, which will be
used for assessing the accuracy versus fairness trade-off. 
\begin{Schunk}
\begin{Sinput}
R> library("ROI.plugin.ecos")
R> library("holiglm")
R> K <- nrow(L)
R> alpha <- seq(0, 1, by = 0.05)
R> pred_prob_race_sex <- sapply(alpha, function(ak) {
+    ck <- ak * s
+    model_constrained <- hglm(as.formula(form), 
+      family = binomial(), data = AdultUCI,
+      scaler = "standardization", big_m = 5,
+      constraints = c(linear(L, rep(">=", K), - ck),
+                      linear(L, rep("<=", K),   ck)))
+    phat <- predict(model_constrained, type = "response")
+    phat
+  })
\end{Sinput}
\end{Schunk}
For each model we compute the in-sample accuracy and area under the ROC curve 
\citep[using package \pkg{pROC},][]{pROC}. Note that we consider a cut-off 
probability threshold of $0.5$, i.e., a label of $\hat y = 1$ will be assigned 
if the predicted probability exceeds 50\%.
To assess the disparate impact of the different models we use the following 
measure:
$$
DI=\frac{\Prob(\hat y = 1|w^{(k)}=0)}{\Prob(\hat y = 1|w^{(k)}=1)}.
$$
We only consider the dummy variable $w^{(k)}$ which corresponds to the 
\code{White:Male} factor combination when computing $DI$ but other approaches 
which take into account all  $w^{(k)}$'s are available. The $DI$ measure
considers the \code{White:Male} as the privileged group in this data set and 
a value less than one indicates that, in the prediction, the privileged group 
is favored. As a rule of thumb, values greater than 0.8 can be considered 
acceptable. \footnote{See also e.g., \href{https://www.eeoc.gov/laws/guidance/questions-and-answers-clarify-and-provide-common-interpretation-uniform-guidelines}{guidelines} of the US Equal Employment Opportunity Commission regarding hiring practices.} 
\begin{Schunk}
\begin{Sinput}
R> obs <- AdultUCI$income
R> res <- apply(pred_prob_race_sex, 2, function(phat) {
+    tab <- table(phat > 0.5, obs)
+    tab_w <- table(phat > 0.5, W[, "sexMale:raceWhite"])
+    ptab_w <- prop.table(tab_w)
+    DI <- ptab_w[2, 1]/ptab_w[2, 2]
+    acc <- sum(diag(tab))/sum(tab)
+    c(DI = DI, acc = acc, auc = pROC::roc(obs, phat, quiet = TRUE)$auc)
+  })
\end{Sinput}
\end{Schunk}
We create the scatterplots of the three measures for the grid of $\alpha$ values:
\begin{Schunk}
\begin{Sinput}
R> par(mfrow = c(1, 3))
R> plot(alpha, res[1,], type = "b",
+       xlab = expression(alpha), ylab = "Disparate impact")
R> plot(alpha, res[2,], type = "b",
+       xlab = expression(alpha), ylab = "Accuracy")
R> plot(alpha, res[3,], type = "b",
+       xlab = expression(alpha), ylab = "AUROC")
\end{Sinput}
\end{Schunk}
The resulting plot can be seen in Figure~\ref{fig:fair:comparison}. We observe 
that the model with $\alpha=0$ has a $DI$ measure of around 0.6, so it still
exhibits disparate impact in favor of the privileged group, even if the 
covariance measure is restricted to zero in the model building phase. To get 
a more accurate picture, one could repeat the analysis by computing the above 
measures on a test set. The cut-off values for the prediction can also be chosen
differently. Finally, one can see that the trade-off between fairness and predictive power 
is not very high, as the loss in the predictive measures is not dramatic as 
$\alpha_k$ decreases.  
Such trade-off analysis can help the modelers and decision-makers in choosing a
setting which is acceptable for the use-case at hand.
\subsection{Model selection in log-binomial regression}
In the following example we show how the \pkg{holiglm} package can be used
for model selection for log-binomial regression.
For this example we use the \code{icu} data set from the \pkg{aplore3}
package~\citep{pkg:aplore3:2016}.  In this application, the goal is to model the 
status of patients at hospital discharge (i.e. alive versus dead) using a series of 
patient-specific covariates.
\begin{Schunk}
\begin{Sinput}
R> data("icu", package = "aplore3")
R> dim(icu)
\end{Sinput}
\begin{Soutput}
[1] 200  21
\end{Soutput}
\end{Schunk}
First, we will check if the MLE exists by making use of the \pkg{detectseparation}
package. For log-binomial regression the existence
criterion is slightly different from the existence criterion for logistic regression, 
for more information we refer to \cite{log-bin:schwendinger:2021}.

\begin{figure}
\centering
\includegraphics[width=\textwidth]{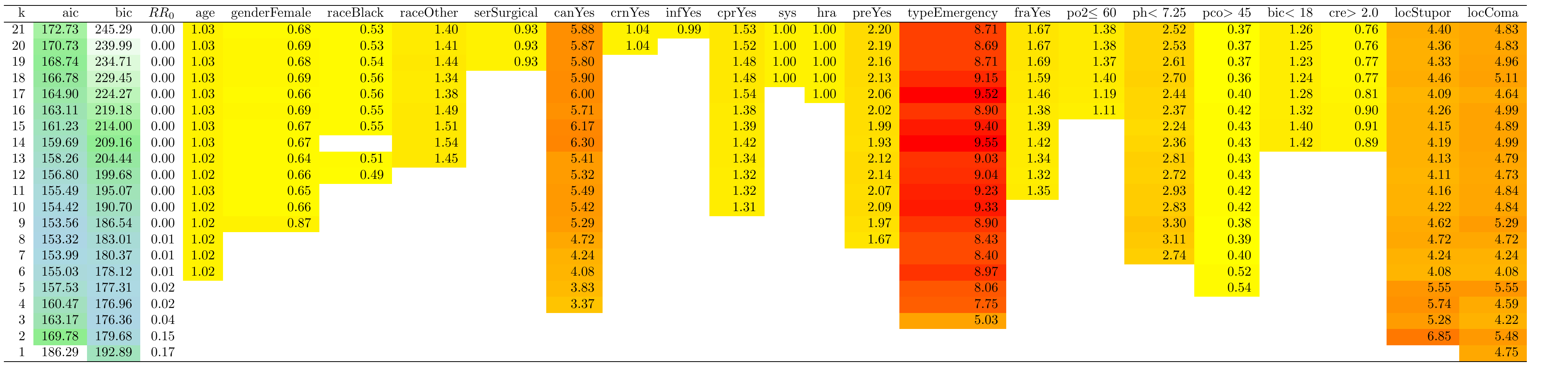}
\caption{AIC, BIC and relative risks ($\exp{\beta_j}$) for the \code{icu} data 
  from the log-binomial
  regression with constraints on the maximum number of covariates $k$. 
  $RR_0$ corresponds to the baseline relative risk $\exp{\beta_0}$.}
 \label{fig:log-binomial}
\end{figure}

\begin{Schunk}
\begin{Sinput}
R> library("detectseparation")
R> icu <- icu[, setdiff(colnames(icu), "id")]
R> glm(sta ~ ., family = binomial("log"), data = icu, 
+    method = "detect_separation")
\end{Sinput}
\begin{Soutput}
Data separation in log-binomial models does not necessarily result in infinite
    estimates
\end{Soutput}
\begin{Soutput}
Implementation: ROI | Solver: lpsolve 
Separation: TRUE 
Existence of maximum likelihood estimates
  (Intercept)           age  genderFemale     raceBlack     raceOther 
            0             0             0             0             0 
  serSurgical        canYes        crnYes        infYes        cprYes 
            0             0             0             0             0 
          sys           hra        preYes typeEmergency        fraYes 
            0             0             0             0             0 
     po2<= 60      ph< 7.25       pco> 45       bic< 18      cre> 2.0 
            0             0             0             0             0 
    locStupor       locComa 
            0             0 
0: finite value, Inf: infinity, -Inf: -infinity
\end{Soutput}
\end{Schunk}
Package \pkg{detectseparation} signals that the data has the separation property;
however, the estimates are still all expected to be finite for the log-binomial
regression. Note that for logistic regression the MLE does not exist 
(i.e. has infinite components). To verify this, again \pkg{detectseparation}
can be used.
\begin{Schunk}
\begin{Sinput}
R> glm(sta ~ ., family = binomial("logit"), data = icu, 
+    method = "detect_separation")
\end{Sinput}
\begin{Soutput}
Implementation: ROI | Solver: lpsolve 
Separation: TRUE 
Existence of maximum likelihood estimates
  (Intercept)           age  genderFemale     raceBlack     raceOther 
            0             0             0          -Inf             0 
  serSurgical        canYes        crnYes        infYes        cprYes 
            0             0             0             0             0 
          sys           hra        preYes typeEmergency        fraYes 
            0             0             0             0             0 
     po2<= 60      ph< 7.25       pco> 45       bic< 18      cre> 2.0 
            0             0             0             0             0 
    locStupor       locComa 
          Inf             0 
0: finite value, Inf: infinity, -Inf: -infinity
\end{Soutput}
\end{Schunk}
For the purpose of model selection, we estimate a sequence of models of 
different size $k=1,\ldots,21$ using function \code{hglm_seq}. 
In Figure~\ref{fig:log-binomial} we present, for each model size $k$, the
AIC and BIC of the optimized model together with the selected coefficients
(which can be interpreted in terms of relative risks for the log-binomial
regression).

\begin{Schunk}
\begin{Sinput}
R> fits <- hglm_seq(formula = sta ~ ., family = binomial("log"),
+    data = icu, solver = "mosek")
\end{Sinput}
\end{Schunk}
In terms of AIC, models of size 7-9 are preferred, while the size 
suggested by BIC is around 3-5. Looking at the estimated set of coefficients for 
each $k$ can provide additional insights into the model behavior.

For example, the modeler can look at the stability of the magnitude of 
the coefficients and how they change when the number of covariates decreases.

%\comnt{LV: we need to include the code for reproducing Figure 2 in the replication material.
%    FS: The code can be found in \code{tables/icu\_table.Rnw} i have not included it
%    since this would add a dependency on \pkg{ztable}, which I would like to avoid.
%}

% ------------------------------------------------------------------------------
%
% Conclusion
%
% ------------------------------------------------------------------------------
\section{Conclusion}\label{sec:conclusions}

The \pkg{holiglm} package can be used to fit generalized linear models 
which are optimal with respect to a set of specified  constraints. The package 
provides different types of constraints which are relevant for statistical 
modeling purposes such as: 
i) different requirements for sparsity 
(e.g. setting an upper limit on the number of covariates included in the model
globally or from a group of pre-specified variables, 
forcing a group of covariates to jointly be included or excluded from the model), 
ii) linear constraints, 
iii) constraints on the pairwise correlation among covariates,
iv) constraints on the sign of the coefficients.
The interface of the package is very similar to the one provided by \code{stats::glm}, 
making it accessible for users to specify the desired GLMs.
Using the fact that the most common GLMs can be represented as conic programs, 
\pkg{holiglm} leverages the tremendous advancements in the field of conic programming 
to provide efficient estimation of constrained GLMs.
This also means that the package will get faster when faster algorithms for conic optimization
will become available.
The package relies on \pkg{ROI} as the underlying infrastructure. Through the 
\pkg{ROI} plugins,  problems with conic objectives and conic- and mixed-integer constraints 
such as the ones implemented in \pkg{holiglm} can reliably be computed within \proglang{R}.
Package \pkg{holiglm} provides an automatic way of estimating the model parameters subject 
to the constraints specified by the modeler. As with any automatic approach to statistical 
modeling,  we recommend to use it complementary to the modeler's expertise as an exploratory 
tool which provides guidance on the structure of the final model.

% ------------------------------------------------------------------------------
%
% Bibs
%
% ------------------------------------------------------------------------------
\bibliography{hglm}

\newpage

% ------------------------------------------------------------------------------
%
% Appendix
%
% ------------------------------------------------------------------------------
\appendix
\section{Conic formulations}\label{sec:formulations}
In this section we reformulate log-likelihood of the different family-link
combinations in terms of Equation~\ref{eq:glmexpfam}.  
Let $\boldsymbol x_{i}^\top \boldsymbol\beta=\eta_i$. 

The family-link combinations implemented in \pkg{holiglm} can be expressed by the 
second-order cone:
$$
\K_\text{soc}^n = \{(t, \boldsymbol x) \in \R^{n}| \boldsymbol x \in \R^{n-1}, t \in \R, \norm{x}_2 \leq t\} 
$$
and the primal exponential cone:
$$
\K_\text{exp} = \{(x, y, z) \in \R^3 | y > 0, y e^{\frac{x}{y}} \leq z\} \cup  \{(x, 0, z) \in \R^3| x \leq 0, z \geq 0\}.
$$
%%%%%%%%%%%%%%%%%%%%%
\subsection{Gaussian}
%%%%%%%%%%%%%%%%%%%%%
For the Gaussian family the log-likelihood is proportional to:
\begin{align*}
\log\mathcal{L}(\boldsymbol\beta; y_1, \dots, y_n)
&\propto\sum_{i=1}^n - \frac{(y_i- g^{-1}(\eta_i))^2}{2}\\
&\propto\sum_{i=1}^n -\frac{y^2_i- 2y_ig^{-1}(\eta_i)+g^{-1}(\eta_i)^2}{2}\propto\sum_{i=1}^n y_i\underbrace{g^{-1}(\eta_i)}_{\lambda_i(\boldsymbol\beta)} - \underbrace{\frac{g^{-1}(\eta_i)^2}{2}}_{b(\lambda_i(\boldsymbol\beta))}
\end{align*}
\subsubsection{Identity link}
For the identity link we have $g^{-1}(\eta_i)=\eta_i$ so 
$$
\lambda_i(\boldsymbol \beta)=\boldsymbol x_{i}^\top \boldsymbol\beta, \qquad
b(\lambda_i(\boldsymbol \beta))=(\boldsymbol x_{i}^\top \boldsymbol\beta)^2/2. 
$$
Using an auxiliary variable $\zeta$, the inequality $\zeta \geq z^2$ can be 
expressed as a second-order cone:
\begin{align*}
%\label{eq:x2quadcone}
\zeta \geq z^2 \iff (\zeta + 1, \zeta - 1, 2 z)\in \K_\text{soc}^{3},
\end{align*}
since 
\begin{align*}
(\zeta + 1, \zeta - 1, 2 z)\in \K_\text{soc}^3 \iff \norm{(\zeta - 1, 2 z)}_2 \leq \zeta + 1
\end{align*}
and by using the definition of the norm and squaring both sides we get
\begin{align*}
(\zeta - 1)^2 + (2 z)^2 \leq (\zeta + 1)^2 \iff 4 z^2 \leq 4 \zeta.
\end{align*}

Using the same technique on $\boldsymbol z \in \R^n$ leads to
\begin{align}
\label{eq:x2quadcone2}
\zeta \geq \sum_{i=1}^n z_i^2 \iff (\zeta + 1, \zeta - 1, 2 z_1, \ldots , 2 z_n)\in \K_\text{soc}^{n+2}.
\end{align}

Since $\text{max} ~ \sum_{i=1}^n y_i \boldsymbol x_i^\top \boldsymbol\beta - \frac{(\boldsymbol x_i^\top \boldsymbol\beta)^2}{2}$
can be rewritten into $\text{max} ~ -\sum_{i=1}^n \frac{1}{2} (y_i - \boldsymbol x_i^\top \boldsymbol\beta)^2 - \frac{y_i^2}{2}$,
and by dropping the constant term $\sum_{i=1}^n \frac{y_i^2}{2}$ we obtain the following reformulation:
\begin{align*}
\underset{\beta, \zeta}{\text{maximize}} ~ & -\zeta \\
\text{subject to} ~ &
( \zeta + 1,
  \zeta - 1,
  2( y_1 -  \boldsymbol x_1^\top \boldsymbol\beta ),
  \ldots, 
  2( y_n -  \boldsymbol x_n^\top\boldsymbol \beta )
) \in \K_\text{soc}^{n+2} \\
& \boldsymbol\beta \in \R^{p}, \zeta \in \R.
\end{align*}

%%%%%%%%%%%%%%%%%%%%%
\subsection{Binomial}
%%%%%%%%%%%%%%%%%%%%%
For the binomial family the log-likelihood is proportional to:
\begin{align*}
\log\mathcal{L}(\boldsymbol\beta; y_1, \dots, y_n)
&\propto \sum_{i=1}^n y_i\log(g^{-1}(\eta_i)) +(1-y_i)\log(1-g^{-1}(\eta_i))\\
&\propto \sum_{i=1}^n y_i\underbrace{(\log(g^{-1}(\eta_i))-\log(1-g^{-1}(\eta_i)))}_{\lambda_i(\boldsymbol\beta)} -
\underbrace{(-\log(1-g^{-1}(\eta_i)))}_{b(\lambda_i(\boldsymbol\beta))}
\end{align*}
\subsubsection{Logit link}
For the logit link we have:
\begin{align*}
\lambda_i(\boldsymbol\beta)&=\log(g^{-1}(\eta_i))-\log(1-g^{-1}(\eta_i)))\\
&=\log\left(\frac{\exp(\eta_i)}{1+\exp(\eta_i)}\right)-\log\left(1-\frac{\exp(\eta_i)}{1+\exp(\eta_i)}\right)=\eta_i\\
b(\lambda_i(\boldsymbol\beta))&=-\log(1-g^{-1}(\eta_i)))\\
&=-\log\left(1-\frac{\exp(\eta_i)}{1+\exp(\eta_i)}\right)=\log\left(1+\exp(\eta_i)\right)\\
\end{align*}

Employing an auxiliary variable $\delta$, the inequality $\log(1+\exp(x))$ can be represented using the exponential cone. 
For introducing the representation, note that the exponential function can be
modeled by an exponential cone:
\begin{align}\label{eq:expcone}
\delta\geq \exp(x) \iff (x, 1, \delta)\in \mathcal{K}_\text{exp}.
\end{align}
Now, the inequality 
\begin{align}
\delta \leq \log(1+\exp(x)) \iff \exp(\delta)\leq 1+\exp(x)
\end{align}
can be reformulated as: 
\begin{align*}
1+\gamma\geq\exp(\delta)&\iff (\delta, 1, 1+\gamma)\in \mathcal{K}_\text{exp}\\
\gamma\geq \exp(x)&\iff (x ,1, \gamma)\in \mathcal{K}_\text{exp}
\end{align*}
The objective is therefore reformulated as:
\begin{align*}
\underset{\boldsymbol\beta,\boldsymbol\delta,\boldsymbol\gamma}{\text{maximize}} & \sum_{i=1}^n y_i\boldsymbol x_{i}^\top \boldsymbol\beta - \delta_i\\
\text{subject to } &  (\delta_i, 1, 1+\gamma_i)\in \mathcal{K}_\text{exp}\text{ for all } i \in \{1, \ldots, n\} \\
& (\boldsymbol x_{i}^\top \boldsymbol\beta, 1, \gamma_i)\in \mathcal{K}_\text{exp}\text{ for all } i \in \{1, \ldots, n\} \\
                 & \boldsymbol\beta \in \R^p, \boldsymbol\delta \in \R^{n},  \boldsymbol\gamma \in \R^{n}.
\end{align*}
\subsubsection{Log link}
For the log link we have:
\begin{align*}
\lambda_i(\boldsymbol\beta)&=\log(g^{-1}(\eta_i))-\log(1-g^{-1}(\eta_i)))\\
&=\log\left(\exp(\eta_i)\right)-\log\left(1-\exp(\eta_i)\right)=\eta_i-\log\left(1-\exp(\eta_i)\right)\\
b(\lambda_i(\boldsymbol\beta))&=-\log(1-g^{-1}(\eta_i)))=-\log\left(1-\exp(\eta_i)\right)
\end{align*}
Similar as before, the inequality $\delta \leq \log(1-\exp(x))$ can be represented using the exponential cone. Now
$$
\delta \leq \log(1-\exp(x)) \iff \exp(\delta) \leq 1-\exp(x)
$$
can be expressed as (using Equation~\ref{eq:expcone}):
\begin{align*}
1-\gamma \geq \exp(\delta) & \iff (\delta,1,1-\gamma)\in \mathcal{K}_\text{exp}\\
\gamma \geq \exp(x) & \iff (x,1,\gamma)\in \mathcal{K}_\text{exp}.
\end{align*}

The objective is therefore reformulated as:
\begin{align*}
\underset{\boldsymbol\beta, \boldsymbol\delta , \boldsymbol\gamma}{\text{maximize}} 
    & \sum_{i=1}^n y_i\boldsymbol x_{i}^\top \boldsymbol\beta + (1-y_i)\delta_i \\
\text{subject to } 
    & \boldsymbol x_{i}^\top \boldsymbol\beta \leq 0 \text{ for all } i \in \{1, \ldots, n\} \\
    & (\delta_i, 1, 1-\gamma_i)\in \mathcal{K}_\text{exp}\text{ for all } i \in \{1, \ldots, n\} \\
    & (\boldsymbol x_{i}^\top \boldsymbol\beta,1,\gamma_i)\in \mathcal{K}_\text{exp}\text{ for all } i \in \{1, \ldots, n\} \\
    & \boldsymbol\beta \in \R^p, \boldsymbol\delta \in \R^{n},  \boldsymbol\gamma \in \R^{n}.
\end{align*}

%%%%%%%%%%%%%%%%%%%%%
\subsection{Poisson}
%%%%%%%%%%%%%%%%%%%%%
For Poisson family the log-likelihood is proportional to:
\begin{align*}
\log\mathcal{L}(\boldsymbol\beta; y_1, \dots, y_n)
&\propto \sum_{i=1}^n  y_i \underbrace{\log(g^{-1}(\eta_i))}_{\lambda_i(\boldsymbol\beta)} -
\underbrace{g^{-1}(\eta_i)}_{b(\lambda_i(\boldsymbol\beta))}
\end{align*}
\subsubsection{Log link}
For the log link we have:
$$
\lambda_i(\boldsymbol\beta)=\eta_i,\qquad
b(\lambda_i(\boldsymbol\beta))=\exp(\eta_i).
$$
Using Equation~\ref{eq:expcone}, the objective can be reformulated as:
\begin{align*}
\underset{\boldsymbol\beta, \boldsymbol\delta}{\text{maximize}} & 
\sum_{i=1}^n y_i\boldsymbol x_{i}^\top \boldsymbol\beta - \delta_i\\
\text{subject to } &  (\boldsymbol x_{i}^\top \boldsymbol\beta, 1, \delta_i)\in \mathcal{K}_\text{exp}\text{ for all } i \in \{1, \ldots, n\} \\
                 & \boldsymbol\beta \in \R^p, \boldsymbol\delta \in \R^{n}.
\end{align*}
\subsubsection{Identity link}
For the identity link we have:
$$
\lambda_i(\boldsymbol\beta)=\log(\eta_i),\qquad
b(\lambda_i(\boldsymbol\beta))=\eta_i.
$$
Imposing  $\eta_i \geq 0$ for all $i=1,\ldots,n$ and using Equation~\ref{eq:expcone}, the objective can be reformulated as:
\begin{align*}
\underset{\boldsymbol\beta, \boldsymbol\delta}{\text{maximize}} & 
\sum_{i=1}^n y_i\delta_i - \boldsymbol x_{i}^\top\boldsymbol\beta \\
\text{subject to } &  (\delta_i, 1, \boldsymbol x_{i}^\top \boldsymbol\beta)\in \mathcal{K}_\text{exp}\text{ for all } i \in \{1, \ldots, n\} \\
                 &\boldsymbol x_{i}^\top \boldsymbol\beta \geq 0 \text{ for all } i \in \{1, \ldots, n\} \\
                 & \boldsymbol\beta \in \R^p, \boldsymbol\delta \in \R^{n}.
\end{align*}
\subsubsection{Square root link}
For the square root link we have:
\begin{align*}
\lambda_i(\boldsymbol\beta)=2\log(\eta_i), \qquad
b(\lambda_i(\boldsymbol\beta))=\eta_i^2
\end{align*}
Using Equations~\ref{eq:x2quadcone2} and ~\ref{eq:expcone}, the objective can be reformulated as:
\begin{align*}
\underset{\boldsymbol\beta, \boldsymbol\delta, \zeta}{\text{maximize}}
    & \sum_{i=1}^n 2y_i \delta_i - \zeta \\
\text{subject to }
    & (\delta_i, 1, \boldsymbol x_{i}^\top \boldsymbol\beta)\in \mathcal{K}_\text{exp}\text{ for all } i \in \{1, \ldots, n\} \\
    & (\zeta + 1,
       \zeta - 1,
       2 \boldsymbol x_{1}^\top \boldsymbol\beta,
       \ldots,
       2 \boldsymbol x_{n}^\top \boldsymbol\beta
      ) \in \mathcal{K}_\textrm{soc}^{n+2} \\
    & \boldsymbol\beta \in \R^p, \boldsymbol\delta \in \R^{n}, \zeta \in \R.
\end{align*}

\end{document}